\pdfoutput=1
\documentclass[letterpaper]{article} 
\usepackage[preprint]{aaai2027}  
\usepackage[hyphens]{url}  
\usepackage{graphicx} 
\usepackage{booktabs}
\usepackage{multirow}
\usepackage{amsmath}
\usepackage{amsthm}
\usepackage{natbib}  
\usepackage{caption} 
\usepackage{amssymb}
\usepackage{algorithm}
\usepackage{xcolor} 
\usepackage{subfigure}
\usepackage{algpseudocode}
\usepackage{newfloat}
\usepackage{listings}
\DeclareCaptionStyle{ruled}{labelfont=normalfont,labelsep=colon,strut=off} 
\floatstyle{ruled}
\newfloat{listing}{tb}{lst}{}
\floatname{listing}{Listing}

\copyrighttext{Open-World Data and Code of EarthLD will be provided once the manuscript is accepted.}

\title{EarthLD: Towards Unified Open-World Landslide Understanding via Vision-Language Guided Diffusion Models}
\author{
    Yuanchao Su\textsuperscript{\rm 1},
    Lianru Gao\textsuperscript{\rm 2}\corresponding,
    Mengying Jiang\textsuperscript{\rm 1},
    Jiangyi Chen\textsuperscript{\rm 3}, \\
    Jiaxin Cheng\textsuperscript{\rm 1},
    and Yicong Zhou\textsuperscript{\rm 1}
}
\affiliations{
    \textsuperscript{\rm 1}Department of Computer and Information Science, University of Macau, Macao 999078, China\\
    \textsuperscript{\rm 2}Key Laboratory of Computational Optical Imaging Technology, Aerospace Information Research Institute, Chinese Academy of Sciences, Beijing 100094, China\\
    \textsuperscript{\rm 3}College of Geomatics, Xi'an University of Science and Technology, Xi'an 710054, China\\
    gaolr@aircas.ac.cn
}

\begin{document}

\maketitle

\begin{abstract}

Landslides are widespread geological hazards, yet their automated detection and mapping in remote sensing imagery remain challenging because of their irregular morphology, ambiguous spectral signatures, and substantial domain shifts across imaging platforms. To overcome these challenges, we propose EarthLD, a vision-language-guided diffusion framework for open-world landslide understanding, enabling unified landslide recognition, mapping, and trigger interpretation. At its core, EarthLD formulates landslide understanding as a diffusion process that progressively infers the presence, spatial extent, and pixel-level boundaries of landslides from noisy latent representations. This probabilistic formulation enables the model to jointly perform image-level landslide recognition and mapping while characterizing predictive uncertainty. By integrating visual observations with contextual knowledge in the denoising process, EarthLD distinguishes diverse landslides from backgrounds, produces confidence-aware predictions for suspected regions, and maps landslide ranges. We additionally construct a global-scale open-world landslide benchmark by systematically harmonizing multiple publicly available remote sensing data collected by diverse institutions. Extensive experiments across regions, sensors, and triggering events demonstrate that EarthLD consistently outperforms existing landslide detection methods, highlighting its potential as a unified and robust solution for global geological-hazard monitoring and emergency response.
\end{abstract}


\section{Introduction}

\begin{figure}[t]
\centering
\includegraphics[height=3.7in]{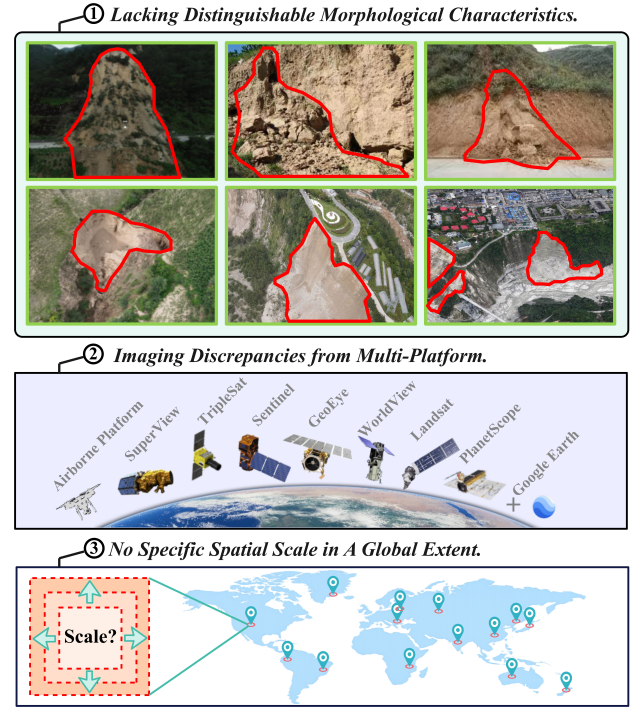}
\caption{Current challenges of global landslide detection.}
\label{teaser}
\end{figure}

Landslides are a gravity-driven geological hazard involving the partial or complete failure of slope materials, in which rock, soil, or their mixtures lose stability and move downslope along a defined or diffuse failure surface~\cite{zhou2025interpretable, chen2024embedding, liu2024research}. The movement may occur in the form of sliding, toppling, falling, or flowing, and can range from slow, progressive deformation to rapid and catastrophic collapse. according to statistics, cause over one thousand deaths worldwide each year and economic losses of billions of US dollars, posing a serious threat to human life and property. Although synthetic aperture radar (SAR), interferometric SAR (InSAR), and ground vibration sensors are valuable for landslide detection, they face practical limitations for large-scale application worldwide. Their operational scale is hindered by fewer satellite platforms and higher data costs compared to optical senors, even when leveraging free data sources like Sentinel-1 and ALOS-2 satellites. Consequently, reliance solely on SAR/InSAR and ground-based sensors is insufficient to meet the demands of global landslide detection. At present, optical sensor-equipped satellites continue to serve as the essential data source for RS imagery in global-scale landslide detection.

Figure.~\ref{teaser} highlights the current main bottleneck for large-scale landslide detection. The specific issues in Figure.~\ref{teaser} are explained as follows:
\textit{1) Unlike conventional object, such as vehicles, airplanes, or buildings, landslides are particularly challenging to identify in multi-platform RS imagery because they have no specific morphological characteristics and distinctive spectral information. 2) Another major difficulty is that imaging cross-sensor introduces fine-grained discrepancies among congeneric objects, including grayscale range, reflectance, and radiance, further compounding the risk of false identification. 3) Even when multispectral or hyperspectral images are adopted, the spectral signatures of landslides are highly similar to those of bare soil, resulting in little to no advantage over common imagery for landslide detection.}

All the time, advances in deep learning and artificial intelligence (AI) have significantly promoted progress in remote sensing image analysis~\cite{ho2020denoising,wyatt2022anoddpm,zhang2024realnet}. More recently, many foundation models were developed to achieve strong contextual understanding and detailed feature representation~\cite{yang2023dawn, li2019visualbert}. They enable learn transferable priors from large-scale data to serve various downstream tasks. Current research on RS foundation models focuses mainly on spatial and spectral feature learning, such as masked autoencoders, SimCLR, SpectralGPT, and SatMAE. Furthermore, some studies have investigated unified interaction mechanisms across spectral, spatial, and frequency domains, such as Alliance. However, these RS foundation models are general-purpose models rather than vertical models specifically designed for landslide detection.

Diffusion models are a specialized class of generative models that learn data distributions through a progressive noising and denoising process~\cite{li2022srdiff}. The most classic example is the Denoising Diffusion Probabilistic Model (DDPM)~\cite{ho2020denoising}. By iteratively transforming random noise into structured representations, they are able to restore missing information and generate high-quality features~\cite{li2022srdiff,kim2024arbitrary}. This iterative refinement endows diffusion models with strong robustness to noise and degraded data, making them particularly effective for object detection in complex or low-quality imagery~\cite{kim2024arbitrary}. Recently, diffusion models have been employed for constructing target detection and semantic segmentation, such as DiffusionDet~\cite{chen2023diffusiondet} and Seg4Diff~\cite{kim2025seg4diff}. Nevertheless, existing methods still face challenges in accurately detecting landslides under conditions like boundary fuzziness, scale variations, or lighting changes~\cite{ho2022video,yang2024fresco}. Moreover, existing diffusion models are not specifically designed to handle the heterogeneity of cross-sensor and multi-resolution data. Their performance tends to degrade under varying radiometric characteristics, acquisition geometries, and resolutions. These limitations suggest that, despite their conceptual advantages, current diffusion-based approaches have yet to fulfill the practical requirements for the robust detection of amorphous and morphologically complex objects like landslides.

To address the aforementioned limitations of existing landslide detection methods, we propose EarthLD, a unified open-world landslide understanding framework based on vision-language guided diffusion models. The main contributions of this work are summarized as follows:

\begin{enumerate}
\item \textbf{Paradigm Shift}: This work re-framed landslide detection as a step-by-step noise reduction process. This new approach makes it much easier to catch landslides with irregular and hard-to-define shapes.
\item \textbf{Model Innovation}: Developed EarthLD, a generative pre-trained foundation model leveraging multi-platform Earth observation data, capable of continuous reinforcement via transfer learning.
\item \textbf{Benchmark Creation}: We curated a global-scale landslide benchmark encompassing 6 continents, 17 countries, and 28 regions to establish a novel unified standard for detection models.
\item \textbf{Practical Impact}: Achieved unprecedented cross-sensor domain generalization, providing a scalable framework for continuous integration of new satellite constellations in large-scale disaster monitoring.
\end{enumerate}

\section{Proposed EarthLD}

\begin{figure*}[t]
\centering
\includegraphics[height=4.7in]{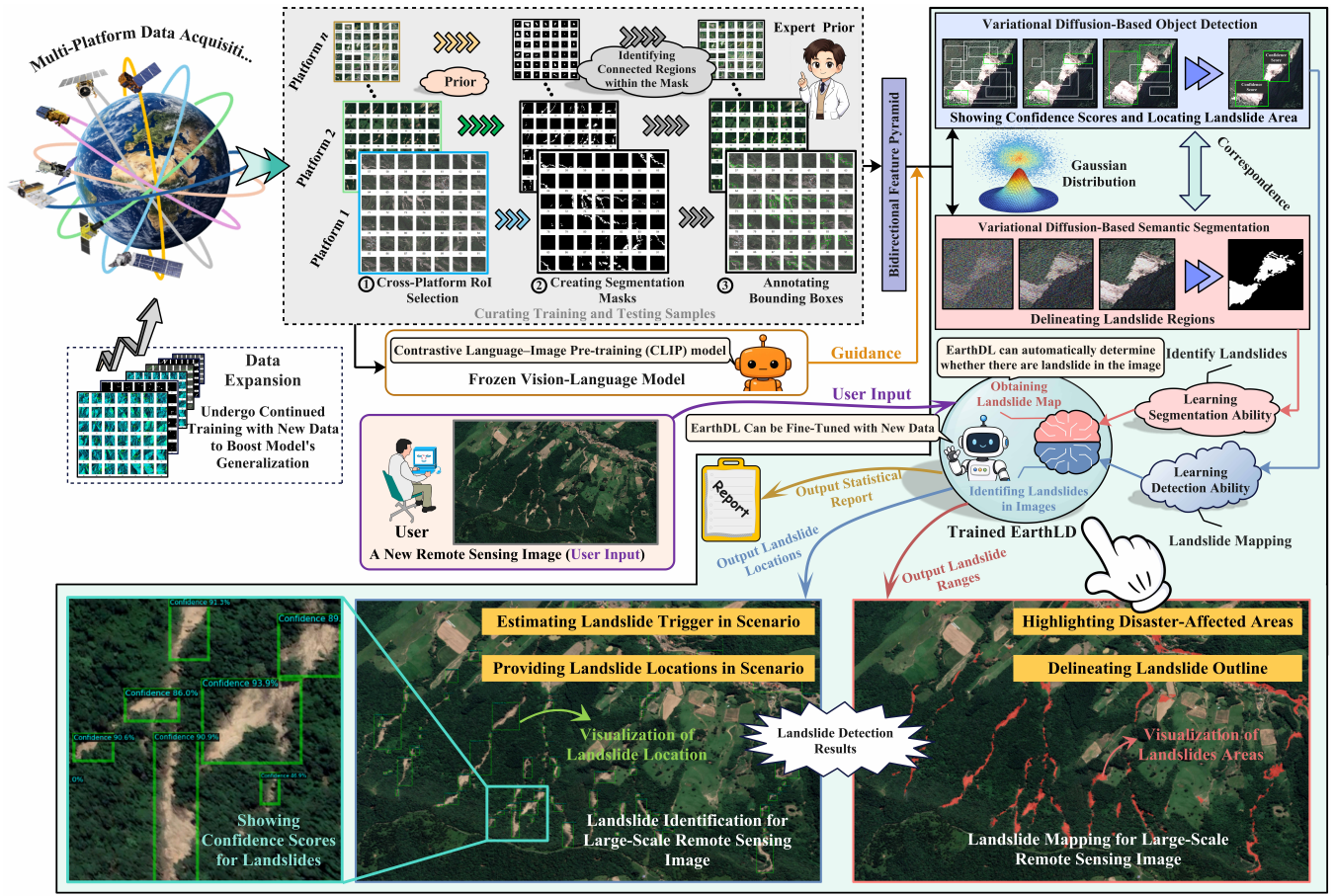}
\caption{Pipeline of EarthLD. Built on a variational diffusion backbone, EarthLD provides unified landslide object detection and semantic segmentation. Once trained, EarthLD accepts a new remote sensing image from any geographic region, visualizes detected landslides, and generates a report summarizing their triggers and total count.}
\label{flowchart}
\end{figure*}

Fig.~\ref{flowchart} illustrates the overall pipeline of EarthLD, an all-in-one framework for landslide recognition, range mapping, trigger estimation, landslide counting, geographic localization. EarthLD realizes \emph{landslide recognition} as an object detection task and \emph{landslide mapping} as a proposal-guided binary semantic segmentation task. 

Specifically, the recognition branch predicts the bounding box and objectness score of each landslide instance, whereas the mapping branch classifies each pixel as landslide foreground or non-landslide background. Thus, a retained detection represents a recognized landslide, and the resulting foreground--background mask represents its mapped spatial extent. EarthLD formulates bounding-box prediction as a Variational Diffusion Model (VDM)~\cite{kingma2021variational} and employs Contrastive Language--Image Pre-training (CLIP)~\cite{radford2021learning} to incorporate temporal, geographical, and trigger-related metadata. A bidirectional feature pyramid network (BiFPN)~\cite{tan2020efficientdet} extracts and refines multi-scale visual representations. The detection branch progressively denoises random box latents into landslide proposals, and the detected boxes are subsequently rasterized as spatial priors for diffusion-based binary segmentation. CLIP-derived metadata representations are fused with both branches to provide contextual guidance.

Given an input remote sensing image $\mathbf{I}\in\mathbb{R}^{H\times W\times C}$ and its metadata $m=(m^{\mathrm{tmp}},m^{\mathrm{geo}},m^{\mathrm{trg}})$, EarthLD predicts a set of recognized landslide instances $\widehat{\mathcal{B}}_{\mathrm{LD}}$, a binary landslide map $\widetilde{\mathbf{M}}$, and an event trigger $\hat{c}$:
\begin{equation}
\mathcal{F}_{\Theta}(\mathbf{I},m)
=
\left(
\widehat{\mathcal{B}}_{\mathrm{LD}},
\widetilde{\mathbf{M}},
\hat{c}
\right).
\label{eq:v2-task}
\end{equation}
The recognition and binary mapping are defined as:
\begin{equation}
\begin{aligned}
\widehat{\mathcal{B}}_{\mathrm{LD}}
&=
\left\{
(\hat{\mathbf{b}}_i,\hat{s}_i)
\mid
\hat{s}_i\geq\delta_{\mathrm{det}}
\right\},\\
\widetilde{\mathbf{M}}
&=
\mathbb{I}\!\left[\hat{\mathbf{M}}\geq\delta_m\right]
\in\{0,1\}^{H\times W},
\end{aligned}
\label{eq:v2-task-output}
\end{equation}
where $\hat{\mathbf{b}}_i=(x_i,y_i,w_i,h_i)\in\mathbb{R}^4$ is the $i$-th predicted box, $\hat{s}_i\in[0,1]$ is its landslide objectness score, and $\delta_{\mathrm{det}}$ is the detection threshold. The matrix $\hat{\mathbf{M}}\in[0,1]^{H\times W}$ is the predicted landslide probability map, $\delta_m$ is the segmentation threshold, and $\mathbb{I}[\cdot]$ denotes the indicator function. Consequently, $|\widehat{\mathcal{B}}_{\mathrm{LD}}|$ gives the detected landslide count, while foreground pixels in $\widetilde{\mathbf{M}}$ delineate the mapped landslide extent. The variable $\hat{c}\in\mathcal{C}_{\mathrm{trigger}}$ denotes the estimated trigger event.

Overall, this work contains three main components: \textit{1) establishing an open-world benchmark for model training and evaluation; 2) developing a CLIP-guided variational diffusion model that realizes landslide recognition through object detection; and 3) developing a proposal-guided variational diffusion model that realizes landslide mapping through binary semantic segmentation.}

\subsection{Establishing Open-World Data Benchmark}
To support the training of EarthLD, this study constructs a comprehensive global-scale landslide dataset by systematically integrating multiple publicly available remote sensing datasets released by different institutions. Instead of relying on a single source, the integrated dataset incorporates heterogeneous remote sensing images from 28 geographically diverse regions across six continents, including Asia, Africa, North America, South America, Europe, and Oceania, covering the period from 2011 to 2022. Such global-scale integration is particularly beneficial for generative modeling because it combines imagery acquired by different sensors and orbital platforms, encompassing diverse imaging geometries, spatial resolutions, noise characteristics, and environmental conditions. By harmonizing these heterogeneous sources into a unified 25.1~GB repository, the proposed dataset enables generative models to learn a comprehensive distribution of global landslide characteristics, ranging from earthquake-induced failures to complex rainfall-triggered landslides.

This study integrates several public landslide datasets, including the GDCLD~\cite{fang2024globally}, Bijie~\cite{ji2020landslide}, CAS~\cite{2024CAS}, GVLM~\cite{zhang2023cross}, and Haiti~\cite{Haiti2010,haidi_sar} datasets. The GDCLD dataset provides more than 15,000 landslide images collected from multiple sensors, including PlanetScope ($3~\mathrm{m}$), GaoFen-6 ($2~\mathrm{m}$), Map World ($0.5~\mathrm{m}$), and UAV imagery ($0.2~\mathrm{m}$). The Bijie dataset contains 2,773 images, including 770 landslide and 2,003 non-landslide samples, collected from TripleSat satellite imagery. The CAS dataset consists of 20,865 image patches acquired from nine study areas using multi-source satellite and UAV imagery.  The GVLM dataset contains 2,895 landslide images collected from Google Earth. The Haiti dataset includes an optical subset with 165 landslide samples derived from post-event GeoEye-1 imagery and pre-event WorldView-2 and Google Earth imagery, as well as a multimodal subset containing 1,713 co-registered SAR--optical image pairs from Sentinel-1 and Sentinel-2. 

The resulting benchmark contains more than 100,000 accurately annotated landslide instances. To ensure cross-sensor consistency, the data were systematically processed through data organization, quality screening, spatial cropping, annotation verification, and conversion into unified detection and segmentation formats.

Let $\mathcal{D}_d$ denote the $d$-th source domain, namely Bijie, Haiti, CAS, GDCLD, or GVLM. If $\mathcal{H}_d$ denotes the corresponding data and annotation harmonization operator, the unified benchmark is
\begin{equation}
\mathcal{D}
=
\bigcup_{d=1}^{5}\mathcal{H}_d(\mathcal{D}_d).
\label{eq:v2-benchmark}
\end{equation}


\subsection{CLIP-Guided Causal Reasoning of Landslides}
To mitigate the spatial ambiguity inherent in monocular landslide detection, we introduce a CLIP-based contextual reasoning module that incorporates auxiliary metadata $m=(m^{\mathrm{tmp}},m^{\mathrm{geo}},m^{\mathrm{trg}})$, representing temporal, geographical, and trigger-related observations, respectively. Here, $m^{\mathrm{trg}}$ contains only observations available at inference time, such as rainfall or seismic measurements and it never contains the ground-truth trigger label.

The metadata are encoded into structured text prompts $\operatorname{Prompt}_r(m^r)$ for $r\in\{\mathrm{tmp},\mathrm{geo},\mathrm{trg}\}$. The prompt feature vectors $\mathbf{t}_r\in\mathbb{R}^D$, where $D$ is the embedding dimension, are computed by the CLIP text encoder $E_T(\cdot)$ and $L_2$-normalized:
\begin{equation}
\mathbf{t}_r
=
\frac{E_T(\operatorname{Prompt}_r(m^r))}
{\|E_T(\operatorname{Prompt}_r(m^r))\|_2}.
\label{eq:v2-meta}
\end{equation}

Let
$\mathbf{e}_I=E_I(\mathbf{I})/\|E_I(\mathbf{I})\|_2\in\mathbb{R}^D$
denote the normalized global visual embedding produced by the CLIP image encoder $E_I(\cdot)$. Cross-modal weights over the three metadata types are computed from cosine similarity:
\begin{equation}
\alpha_r
=
\frac{\exp(\mathbf{e}_I^\top\mathbf{t}_r/\tau)}
{\sum_{q\in\{\mathrm{tmp},\mathrm{geo},\mathrm{trg}\}}
\exp(\mathbf{e}_I^\top\mathbf{t}_q/\tau)},
\label{eq:v2-meta-attn}
\end{equation}
where $\tau\in\mathbb{R}^{+}$ is a learnable temperature. The aggregated metadata representation is
\begin{equation}
\mathbf{q}_m
=
\mathbf{W}_m
\left(
\sum_r\alpha_r\mathbf{t}_r
\right),
\label{eq:v2-meta-fuse}
\end{equation}
where $\mathbf{W}_m\in\mathbb{R}^{D\times D}$ is a learnable projection matrix.

\subsection{CLIP-Guided Variational Diffusion Model for Landslide Recognition}
EarthLD realizes landslide recognition through class-agnostic object detection: every retained object corresponds to one recognized landslide instance. Therefore, the recognition branch predicts landslide bounding boxes and their objectness scores. 

A backbone network extracts a multi-scale feature pyramid $\{\mathbf{P}_k\}_{k=1}^{K}$, where $\mathbf{P}_k\in\mathbb{R}^{H_k\times W_k\times C_k}$. BiFPN aggregates adjacent feature levels into refined features $\mathbf{F}_k\in\mathbb{R}^{H_k\times W_k\times C}$:
\begin{equation}
\begin{aligned}
\mathbf{F}_k
&=
\operatorname{Conv}
\left(
\sum_{j\in\mathcal{N}(k)}
\omega_{kj}\operatorname{Resize}(\mathbf{P}_j,k)
\right),\\
\omega_{kj}
&=
\frac{\operatorname{ReLU}(w_{kj})}
{\varepsilon+\sum_{r\in\mathcal{N}(k)}
\operatorname{ReLU}(w_{kr})},
\end{aligned}
\label{eq:v2-weight}
\end{equation}
where $\mathcal{N}(k)$ is the set of levels connected to level $k$, $w_{kj}\in\mathbb{R}$ are learnable scalar weights, and $\varepsilon>0$ prevents division by zero. The fused features are flattened into a multi-scale visual token matrix
\begin{equation}
\mathbf{V}
=
\operatorname{Flatten}(\mathbf{F}_1,\ldots,\mathbf{F}_K),
\quad
L=\sum_{k=1}^{K}H_kW_k.
\label{eq:v2-visual}
\end{equation}

For detection query $i$, let $\mathbf{q}_i$ be a learnable query vector. Its proposal-wise condition is defined by
\begin{equation}
\mathbf{c}_i
=
\mathbf{W}_c
\left[
\operatorname{Attn}(\mathbf{q}_i,\mathbf{V})
\,\|\,\mathbf{q}_m
\right],
\label{eq:v2-box-cond}
\end{equation}
where $\operatorname{Attn}(\cdot)$ retrieves query-specific visual evidence, $\|$ denotes concatenation, and $\mathbf{W}_c$ projects the fused visual-metadata representation.

\subsubsection{Variational Diffusion-Based Landslide Object Detection:}

For a ground-truth box $\mathbf{b}\in\mathbb{R}^4$, the variational encoder models the clean box latent $\mathbf{z}_0^{\mathbf{b}}$ conditioned on $\mathbf{c}_i$:
\begin{equation}
q_\phi(\mathbf{z}_0^{\mathbf{b}}\mid\mathbf{b},\mathbf{c}_i) =
\mathcal{N}
\left(
\mathbf{z}_0^{\mathbf{b}};
\boldsymbol{\mu}_\phi(\mathbf{b},\mathbf{c}_i),
\operatorname{diag}
\left(
(\boldsymbol{\sigma}_\phi(\mathbf{b},\mathbf{c}_i))^2
\right)
\right).
\label{eq:v2-box-forward}
\end{equation}
where $\boldsymbol{\mu}_\phi(\cdot), \boldsymbol{\sigma}_\phi(\cdot)$ are the predicted mean and standard deviation vectors.
For a continuous diffusion time $t\in[0,1]$, a learnable log-SNR schedule $\gamma_\nu(t)$ parameterizes the forward transition:
\begin{equation}
q_\nu(\mathbf{z}_t^{\mathbf{b}}\mid\mathbf{z}_0^{\mathbf{b}}) =
\mathcal{N}
\left(
\mathbf{z}_t^{\mathbf{b}};
\alpha_t\mathbf{z}_0^{\mathbf{b}},
\sigma_t^2\mathbf{I}_4
\right),
\label{eq:v2-box-noise}
\end{equation}
where $\mathbf{I}_4\in\mathbb{R}^{4\times4}$ is the identity matrix, $\alpha_t^2=\operatorname{sigmoid}(-\gamma_\nu(t))$, and $\sigma_t^2=\operatorname{sigmoid}(\gamma_\nu(t))$.

For $0\leq s<t\leq1$, the parameterized reverse transition is
\begin{equation}
p_\theta
(\mathbf{z}_s^{\mathbf{b}}
\mid\mathbf{z}_t^{\mathbf{b}},\mathbf{c}_i)
=
\mathcal{N}
\big(
\mathbf{z}_s^{\mathbf{b}};
\boldsymbol{\mu}_\theta
(\mathbf{z}_t^{\mathbf{b}},s,t,\mathbf{c}_i),
\boldsymbol{\Sigma}_\theta
(\mathbf{z}_t^{\mathbf{b}},s,t,\mathbf{c}_i)
\big).
\label{eq:v2-box-clean}
\end{equation}

The Evidence Lower Bound (ELBO) objective for box diffusion is
\begin{equation}
\begin{aligned}
\mathcal{L}_{\mathrm{VDM}}^{\mathbf{b}}
={}&
D_{\mathrm{KL}}
\left(
q_{\phi,\nu}
(\mathbf{z}_1^{\mathbf{b}}\mid\mathbf{b},\mathbf{c}_i)
\|
p(\mathbf{z}_1^{\mathbf{b}})
\right)\\
&-
\mathbb{E}_{q_\phi}
\left[
\log p_\kappa
(\mathbf{b}\mid\mathbf{z}_0^{\mathbf{b}},\mathbf{c}_i)
\right]
+
\int_0^1
\mathcal{L}_{\mathrm{diff}}^{\mathbf{b}}(t)\,\mathrm{d}t,
\end{aligned}
\label{eq:v2-box-reverse}
\end{equation}
where $D_{\mathrm{KL}}(\cdot \,||\, \cdot)$ is the KL divergence scalar, $q_{\phi,\nu}
(\mathbf{z}_1^{\mathbf{b}}\mid\mathbf{b},\mathbf{c}_i)
= \int q_\nu(\mathbf{z}_1^{\mathbf{b}}\mid\mathbf{z}_0^{\mathbf{b}})
q_\phi(\mathbf{z}_0^{\mathbf{b}}\mid\mathbf{b},\mathbf{c}_i)
\,\mathrm{d}\mathbf{z}_0^{\mathbf{b}}$, and $p(\mathbf{z}_1^{\mathbf{b}})=\mathcal{N}(\mathbf{0},\mathbf{I}_4)$.

The decoded clean latent yields a box and a landslide object score:
\begin{equation}
(\hat{\mathbf{b}}_i,\hat{s}_i)
=
D_\kappa^b
(\hat{\mathbf{z}}_{0,i}^{\mathbf{b}},\mathbf{c}_i).
\label{eq:v2-box-mean}
\end{equation}
As defined in Eq.~\eqref{eq:v2-task-output}, boxes satisfying $\hat{s}_i\geq\delta_{\mathrm{det}}$ form $\widehat{\mathcal{B}}_{\mathrm{LD}}$. Hence, object detection directly produces the landslide recognition result, and the number of retained boxes gives the landslide count. 

Algorithm~\ref{alg:laddetection} shows the pseudocode for variational diffusion-based landslide object detection.

\begin{algorithm}[t]
\caption{Variational Diffusion-Based Landslide Object Detection}
\begin{algorithmic}[1]
\Require Image $\mathbf{I}$, Metadata $m$, Query $\mathbf{q}_i$, Threshold $\delta_{\mathrm{det}}$
\Ensure Landslides $\widehat{\mathcal{B}}_{\mathrm{LD}}$
\State $\mathbf{q}_m \leftarrow \text{CLIP}(m, \mathbf{I})$, \quad $\mathbf{V} \leftarrow \text{BiFPN}(\mathbf{I})$ using~{Eq.~\eqref{eq:v2-meta}--\eqref{eq:v2-visual}}
\State $\mathbf{c}_i \leftarrow \mathbf{W}_c [\operatorname{Attn}(\mathbf{q}_i, \mathbf{V}) \,\|\, \mathbf{q}_m]$ using~{Eq.~\eqref{eq:v2-box-cond}}
\State $\mathbf{z}_0^{\mathbf{b}} \sim q_\phi(\mathbf{z}_0^{\mathbf{b}} \mid \mathbf{b}, \mathbf{c}_i)$, \quad $\mathbf{z}_t^{\mathbf{b}} \sim q_\nu(\mathbf{z}_t^{\mathbf{b}} \mid \mathbf{z}_0^{\mathbf{b}})$ using~{Eq.~\eqref{eq:v2-box-forward}--\eqref{eq:v2-box-noise}}
\State Optimize via $\mathcal{L}_{\mathrm{VDM}}^{\mathbf{b}}$ using~{Eq.~\eqref{eq:v2-box-reverse}}
\State $\mathbf{z}_1^{\mathbf{b}} \sim \mathcal{N}(\mathbf{0}, \mathbf{I}_4)$ \\
\For{$t = 1 \text{ down to } 0$}
\State $\mathbf{z}_s^{\mathbf{b}} \sim p_\theta(\mathbf{z}_s^{\mathbf{b}} \mid \mathbf{z}_t^{\mathbf{b}}, \mathbf{c}_i)$ \Comment{Denoise via Eq.~\eqref{eq:v2-box-clean}}
\State $(\hat{\mathbf{b}}_i, \hat{s}_i) \leftarrow D_\kappa^b(\mathbf{z}_0^{\mathbf{b}}, \mathbf{c}_i)$ using~{Decode via Eq.~\eqref{eq:v2-box-mean}}
\EndFor
\State \Return $\widehat{\mathcal{B}}_{\mathrm{LD}} = \{ (\hat{\mathbf{b}}_i, \hat{s}_i) \mid \hat{s}_i \geq \delta_{\mathrm{det}} \}$ using~{Eq.~\eqref{eq:v2-task-output}}
\end{algorithmic}
\label{alg:laddetection}
\end{algorithm}

\subsubsection{Auxiliary Open-World and Trigger Interpretation:}
The preceding detection output is sufficient to define landslide recognition. After detection, an auxiliary interpretation module may additionally assign semantic labels to recognized instances and estimate their likely trigger; these auxiliary predictions do not change the definition of recognition as object detection.

Let $\mathbf{h}_i=\mathbf{f}_i/\|\mathbf{f}_i\|_2$ denote the normalized embedding of detected proposal $i$, and let $\mathbf{p}_k$ be the normalized text prototype for semantic category $k\in\{1,\ldots,K_c\}$. The posterior probability over known categories is
\begin{equation}
\pi_{ik}
=
\frac{\exp(\mathbf{h}_i^\top\mathbf{p}_k/\tau_o)}
{\sum_{j=1}^{K_c}
\exp(\mathbf{h}_i^\top\mathbf{p}_j/\tau_o)},
\label{eq:v2-prototype}
\end{equation}
where $\tau_o\in\mathbb{R}^{+}$ is a temperature. The energy-based uncertainty score is
\begin{equation}
u_i =
-\tau_o
\log
\sum_{k=1}^{K_c}
\exp
\left(
\frac{\mathbf{h}_i^\top\mathbf{p}_k}{\tau_o}
\right).
\label{eq:v2-energy}
\end{equation}
Open-world semantic assignment is then given by
\begin{equation}
\hat{y}_i =
\begin{cases}
\arg\max_k\pi_{ik},
&\text{if }
\max_k(\mathbf{h}_i^\top\mathbf{p}_k)\geq\delta_s
\text{ and }u_i\leq\delta_u,\\
\mathrm{Unknown},
&\text{otherwise}.
\end{cases}
\label{eq:v2-reject}
\end{equation}

For trigger estimation, each candidate trigger $c\in\mathcal{C}_{\mathrm{trigger}}$ is mapped to a normalized text prototype $\mathbf{t}_c$. Global image evidence and recognized-instance evidence are combined as:
\begin{equation}
\rho_c =
\begin{cases}
\lambda_I(\mathbf{e}_I^\top\mathbf{t}_c)
+
(1-\lambda_I)
\dfrac{1}{N}
\displaystyle\sum_{i=1}^{N}
(\mathbf{h}_i^\top\mathbf{t}_c),
&N>0,\\[6pt]
\mathbf{e}_I^\top\mathbf{t}_c,
&N=0,
\end{cases}
\label{eq:v2-trigger}
\end{equation}
where $\lambda_I\in[0,1]$ balances global and instance-level evidence. The second branch avoids division by zero when no detection survives filtering. The predicted trigger is
\begin{equation}
\hat{c}
=
\arg\max_{c\in\mathcal{C}_{\mathrm{trigger}}}\rho_c.
\label{eq:v2-trigger-output}
\end{equation}

\subsection{Proposal-Guided Variational Diffusion for Binary Semantic Segmentation (Landslide Mapping)}
EarthLD realizes landslide mapping as binary semantic segmentation. Each pixel is assigned to either the landslide foreground ($1$) or the non-landslide background ($0$), while the recognized bounding boxes provide spatial proposals that guide the delineation of irregular landslide boundaries.

Let $\mathbf{M}\in\{0,1\}^{H\times W}$ denote the ground-truth binary mask. Its variational encoder models a clean spatial latent $\mathbf{z}_0^{\mathbf{M}}\in\mathbb{R}^{H'\times W'\times C_m}$:
\begin{equation}
q_\varphi
(\mathbf{z}_0^{\mathbf{M}}\mid\mathbf{M},\mathbf{I})
=
\mathcal{N}
\big(
\mathbf{z}_0^{\mathbf{M}};
\boldsymbol{\mu}_\varphi^{\mathbf{M}}(\mathbf{M},\mathbf{I}),
\operatorname{diag}
\big(
(\boldsymbol{\sigma}_\varphi^{\mathbf{M}}
(\mathbf{M},\mathbf{I}))^2
\big)
\big).
\label{eq:v2-mask-forward}
\end{equation}

The mask latent follows the same log-SNR forward process as the box latent:
\begin{equation}
q_\nu
(\mathbf{z}_t^{\mathbf{M}}\mid\mathbf{z}_0^{\mathbf{M}})
=
\mathcal{N}
\left(
\mathbf{z}_t^{\mathbf{M}};
\alpha_t\mathbf{z}_0^{\mathbf{M}},
\sigma_t^2\mathbf{I}_{M}
\right),
\label{eq:v2-mask-noise}
\end{equation}
where $\mathbf{I}_{M}$ is the identity operator over the mask latent.

The recognized boxes $\widehat{\mathcal{B}}_{\mathrm{LD}}$ are rasterized into a spatial proposal map $\mathbf{S}_B\in[0,1]^{H\times W}$. The conditioning tensor fuses multi-scale spatial features, box proposals, and CLIP-derived metadata:
\begin{equation}
\mathbf{x}_M
=
\operatorname{Conv}
\left(
\left[
\operatorname{Up}(\mathbf{F}_{1:K})
\,\|\,\mathbf{S}_B
\,\|\,\operatorname{Tile}(\mathbf{q}_m)
\right]
\right),
\label{eq:v2-mask-cond}
\end{equation}
where $\operatorname{Up}(\mathbf{F}_{1:K})$ upsamples and concatenates the feature maps at resolution $H\times W$, and $\operatorname{Tile}(\mathbf{q}_m)$ spatially replicates $\mathbf{q}_m$ over the same grid.

The segmentation denoiser estimates the clean mask latent, which is decoded into a landslide probability map:
\begin{equation}
\hat{\mathbf{z}}_0^{\mathbf{M}}
=
G_\psi(\mathbf{z}_t^{\mathbf{M}},t,\mathbf{x}_M), \qquad
\hat{\mathbf{M}}
=
\operatorname{sigmoid}
\left(
D_\kappa^M
(\hat{\mathbf{z}}_0^{\mathbf{M}},\mathbf{x}_M)
\right).
\label{eq:v2-mask-clean}
\end{equation}
The final binary landslide map is
\begin{equation}
\widetilde{\mathbf{M}}_{h,w}
=
\begin{cases}
1,&\hat{\mathbf{M}}_{h,w}\geq\delta_m,\\
0,&\hat{\mathbf{M}}_{h,w}<\delta_m.
\end{cases}
\label{eq:v2-binary-map}
\end{equation}
Therefore, binary semantic segmentation directly produces the landslide mapping result: pixels labeled $1$ delineate landslide regions, whereas pixels labeled $0$ represent the background.

The mask diffusion loss is
\begin{equation}
\mathcal{L}_{\mathrm{VDM}}^{\mathbf{M}}
=
\mathbb{E}_{t,\mathbf{z}_0^{\mathbf{M}},\boldsymbol{\epsilon}}
\left[
w(t)
\left\|
\boldsymbol{\epsilon}
-
\boldsymbol{\epsilon}_\psi
(\mathbf{z}_t^{\mathbf{M}},t,\mathbf{x}_M)
\right\|_2^2
\right],
\label{eq:v2-mask-vdm}
\end{equation}
where $\boldsymbol{\epsilon}\sim\mathcal{N}(\mathbf{0},\mathbf{I}_{M})$ and $w(t)\geq0$ is the weighting induced by the selected log-SNR parameterization. 

Algorithm~\ref{alg:mapping} provides the pseudocode of the binary semantic segmentation.

\begin{algorithm}[t]
\caption{Proposal-Guided Diffusion for Landslide Mapping}
\label{alg:mapping}
\begin{algorithmic}[1]
\Require Features $\mathbf F_{1:K}$, detected boxes
         $\widehat{\mathcal B}$, metadata $\mathbf q_m$, GT mask $\mathbf M$
\Ensure Binary landslide mask $\widetilde{\mathbf M}$
\State Rasterize $\widehat{\mathcal B}$ into proposal prior $\mathbf S_B$
\State Construct mask condition $\mathbf x_M$ using~{\eqref{eq:v2-mask-cond}}
\State Encode and diffuse $\mathbf M$ into $\mathbf z_t^{\mathbf M}$ using~{\eqref{eq:v2-mask-forward}, \eqref{eq:v2-mask-noise}}
\State $\widehat{\mathbf z}_0^{\mathbf M}
       \gets G_\psi(\mathbf z_t^{\mathbf M},t,\mathbf x_M)$ using~{\eqref{eq:v2-mask-clean}}
\State $\widehat{\mathbf M}\gets
       \sigma(D_\kappa^M(\widehat{\mathbf z}_0^{\mathbf M},\mathbf x_M))$ using~{\eqref{eq:v2-mask-clean}}
\State Optimize $\mathcal L_{\mathrm{map}}$ using~{\eqref{eq:v2-mask-clean}, \eqref{eq:v2-mask-vdm}, \eqref{eq:v2-map-loss}}
\State \Return $\widetilde{\mathbf M}$
\end{algorithmic}
\end{algorithm}

\begin{figure}[t]\scriptsize
\begin{tabular}{p{3.7cm}p{3.7cm}}
\includegraphics[height=1.5in]{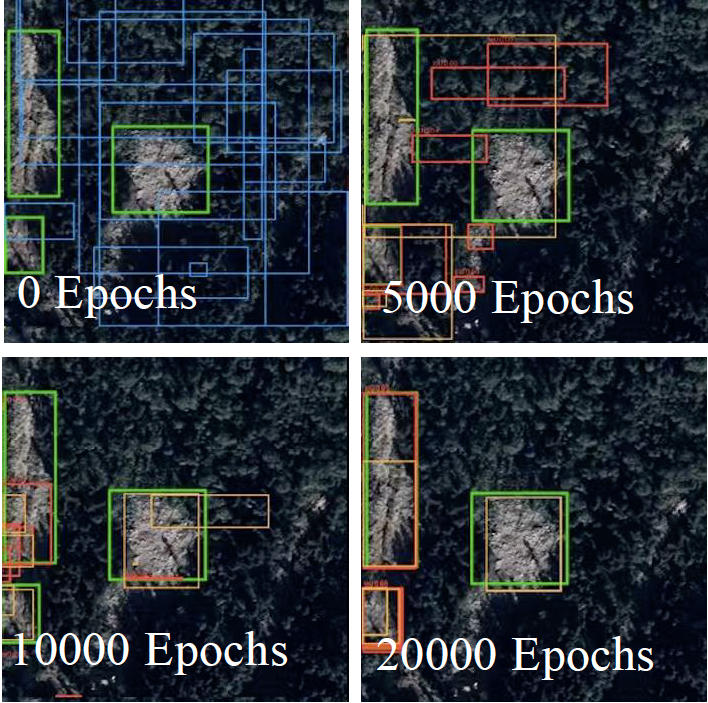}&
\includegraphics[height=1.5in]{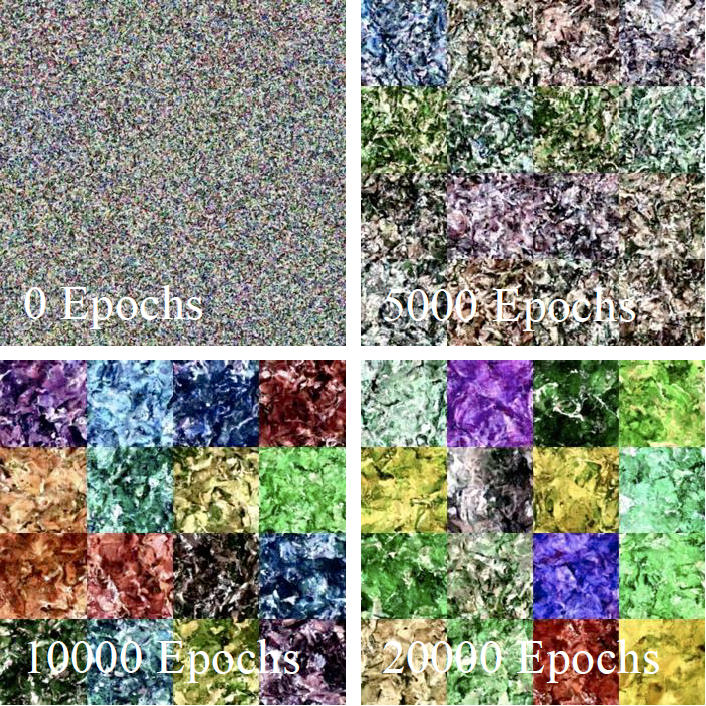}\\
\hspace{1.0cm}Detection Learning  &\hspace{0.7cm}Segmentation Learning \\
\end{tabular}
\caption{The progressive improvement in feature representation for both object detection and semantic segmentation by the training, as observed from the qualitative outputs at corresponding epochs.}
\label{Middle_showing}
\end{figure}

\begin{figure}[t]\scriptsize
\begin{tabular}{p{1.65cm}p{1.65cm}p{1.65cm}p{1.65cm}}
\includegraphics[height=1.5in]{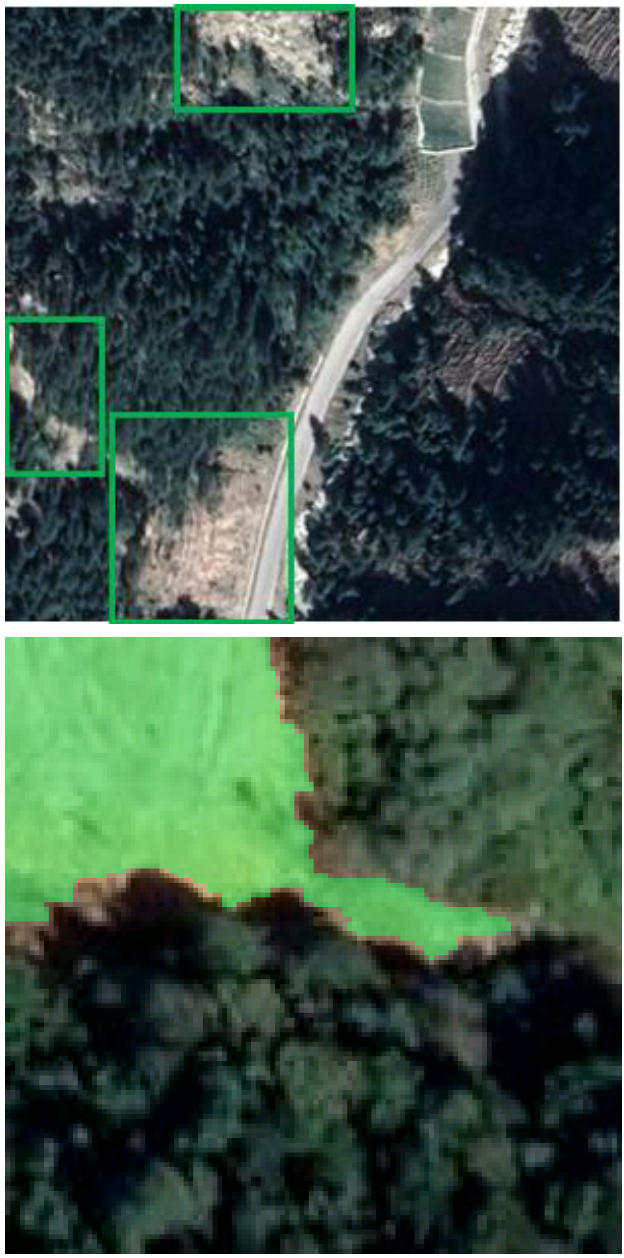}&
\includegraphics[height=1.5in]{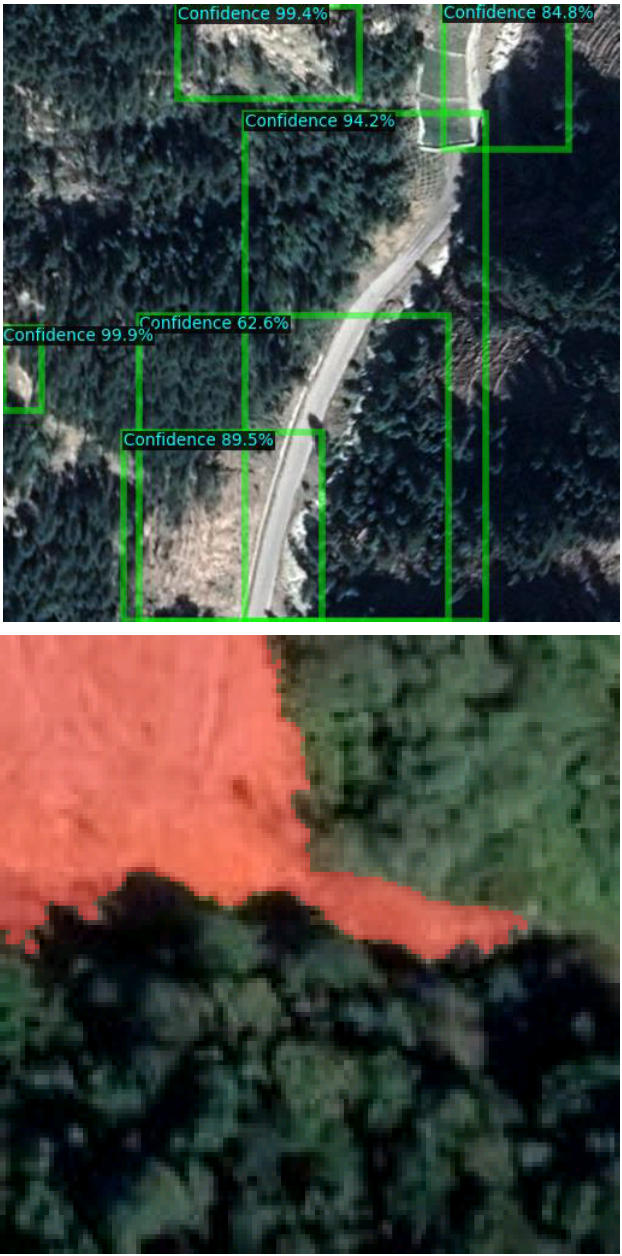}&
\includegraphics[height=1.5in]{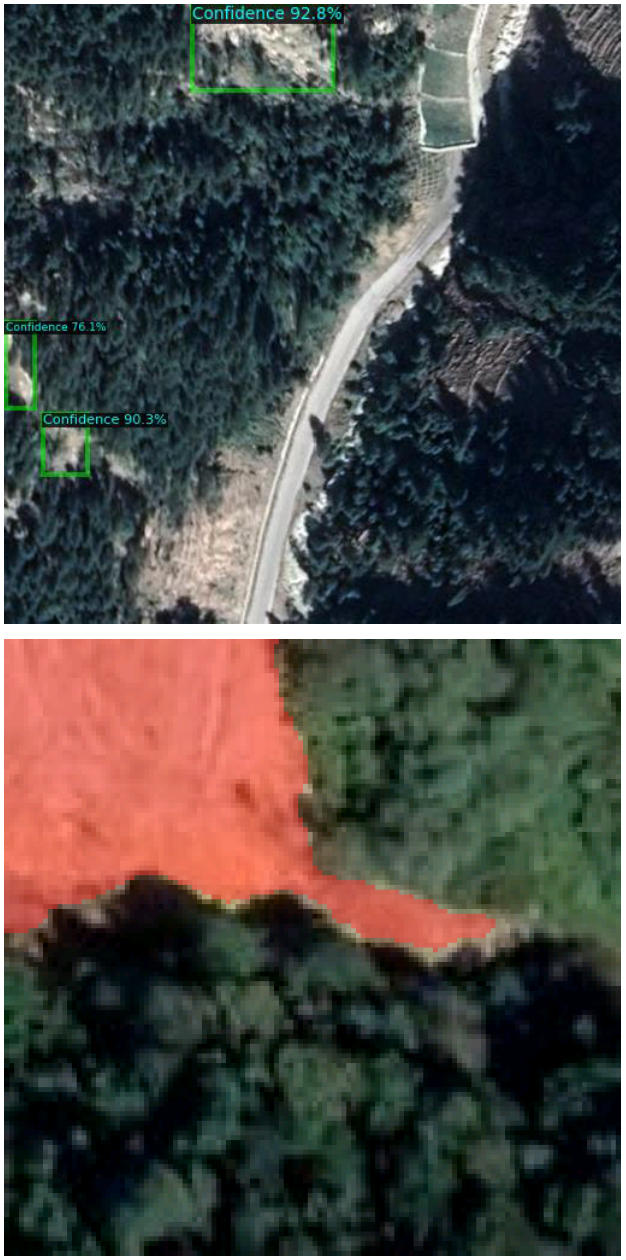}&
\includegraphics[height=1.5in]{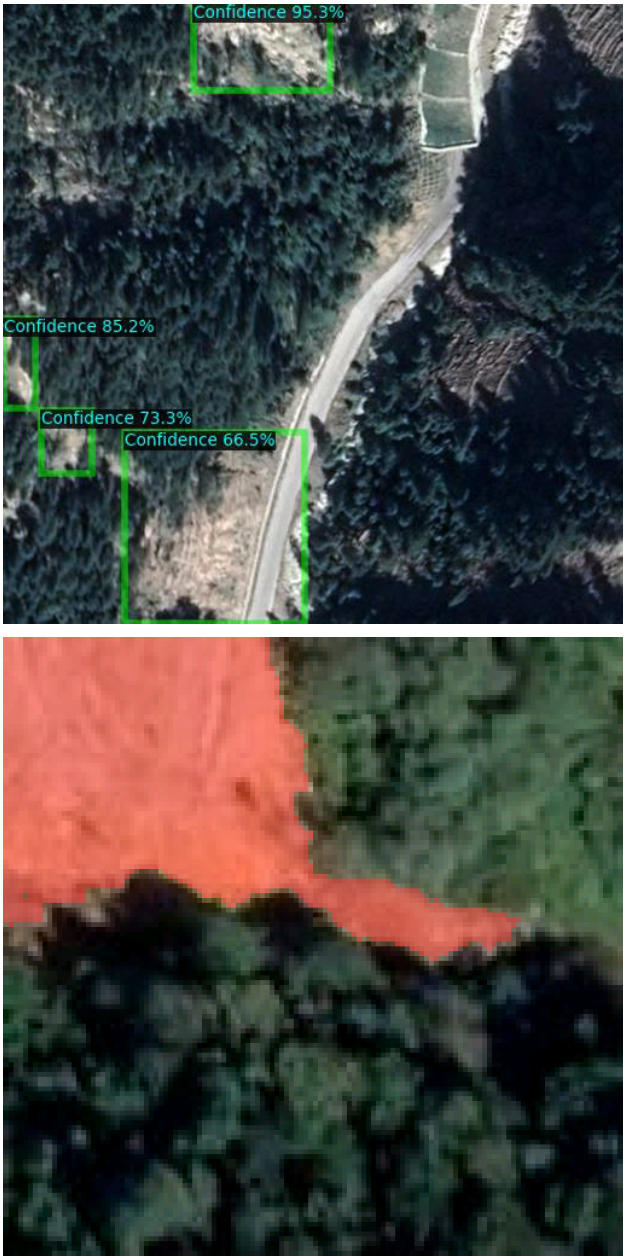}\\
\hspace{0.8cm}GT &\hspace{0.5cm} No CLIP &\hspace{0.3cm}No Pre-Train &\hspace{0.5cm}EarthLD \\
\end{tabular}
\caption{Ablation study of CLIP guidance and pre-training on EarthLD. The full EarthLD model achieves the best performance, whereas removing CLIP guidance or pre-training leads to a noticeable degradation in landslide identification and mapping.}
\label{pre-traintest}
\end{figure}

\begin{figure}[t]\scriptsize
\centering
\begin{tabular}{p{1.6cm}p{1.6cm}p{1.6cm}p{1.6cm}}
\includegraphics[height=1.3in]{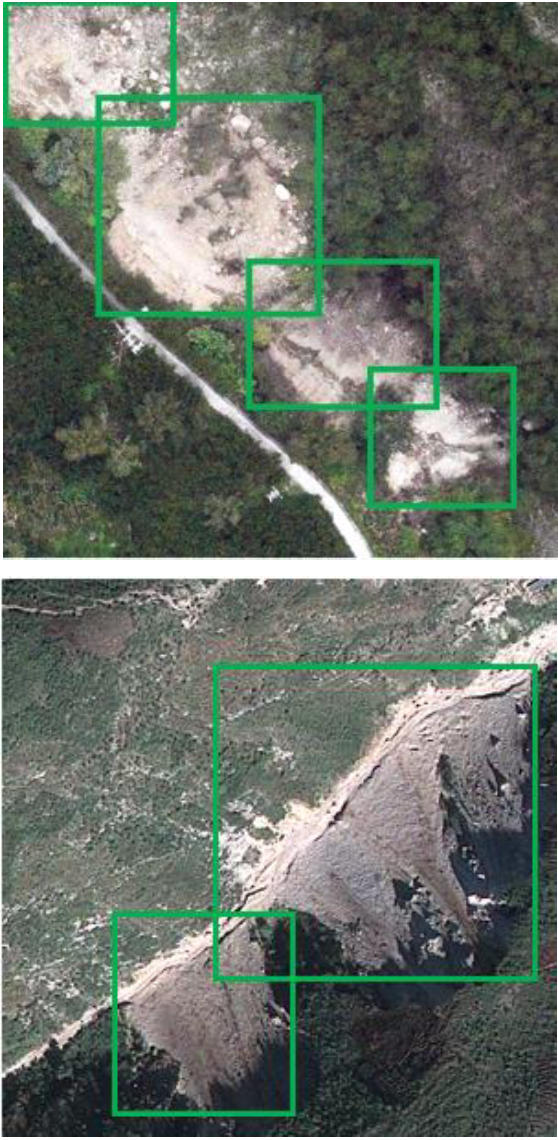}&
\includegraphics[height=1.3in]{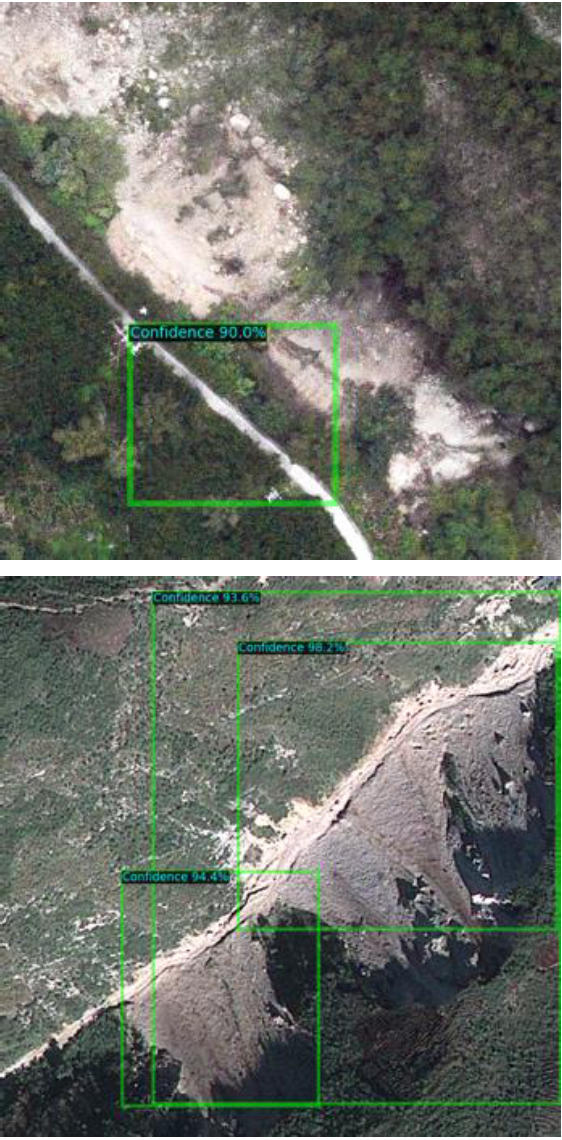}&
\includegraphics[height=1.3in]{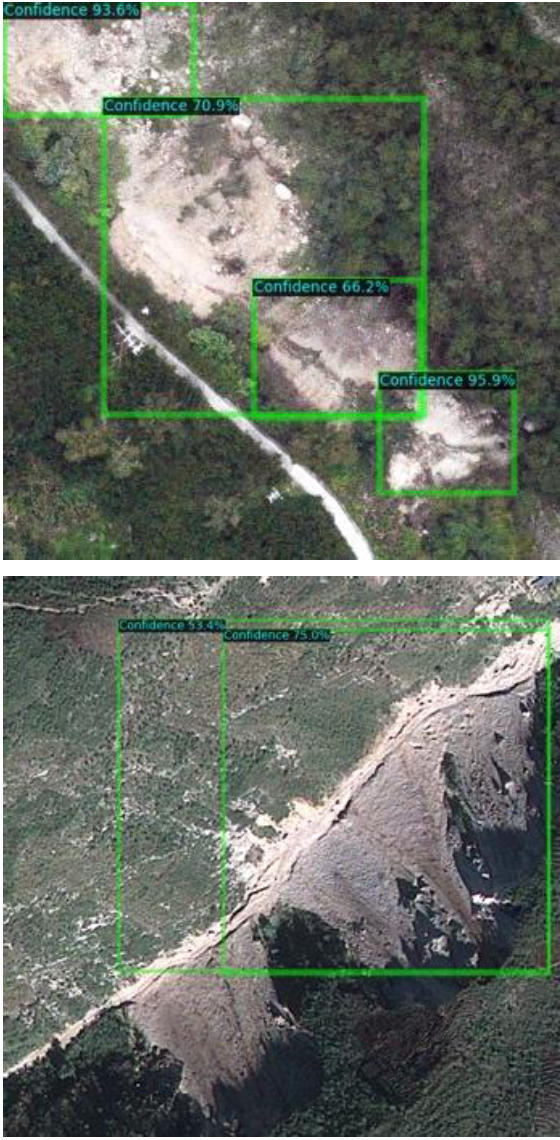}&
\includegraphics[height=1.3in]{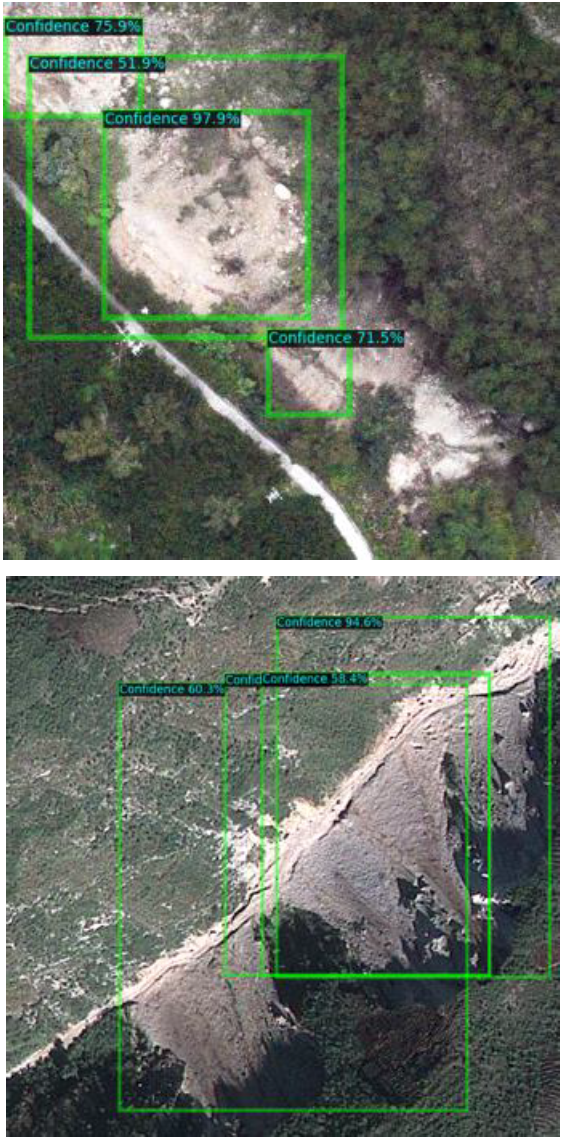}\\
\hspace{0.6cm}GT &\hspace{0.1cm}Faster R-CNN&\hspace{0.1cm}Sparse R-CNN &\hspace{0.2cm}DAB-DETR \\
\includegraphics[height=1.3in]{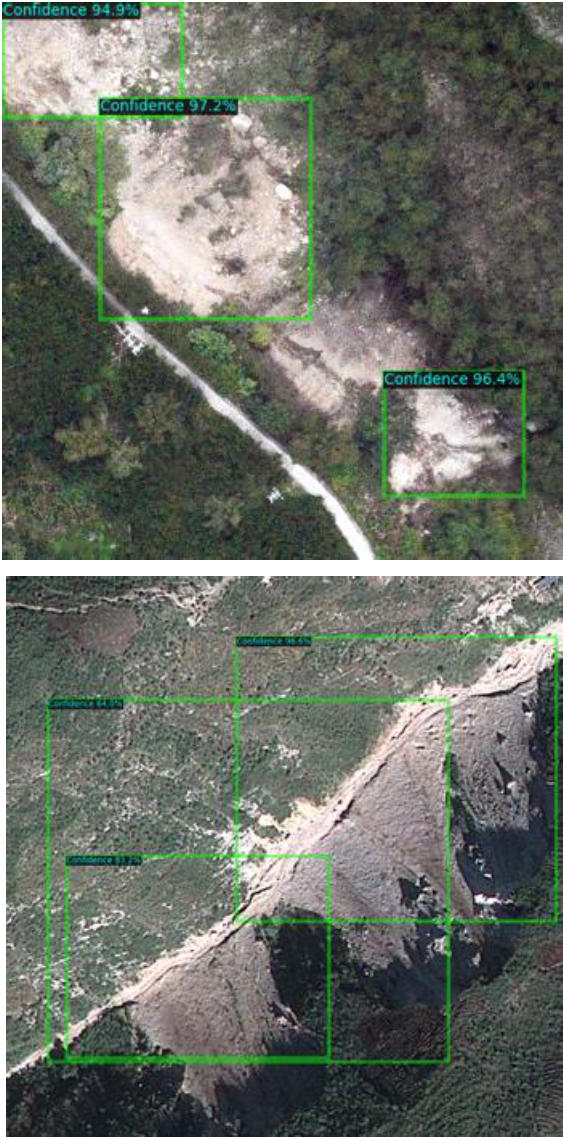}&
\includegraphics[height=1.3in]{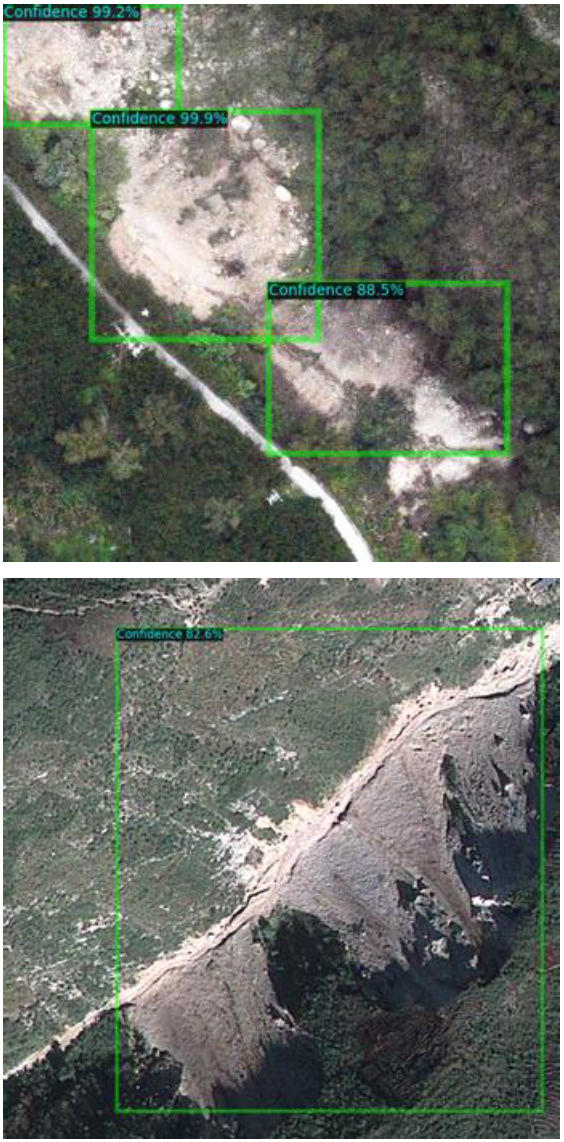}&
\includegraphics[height=1.3in]{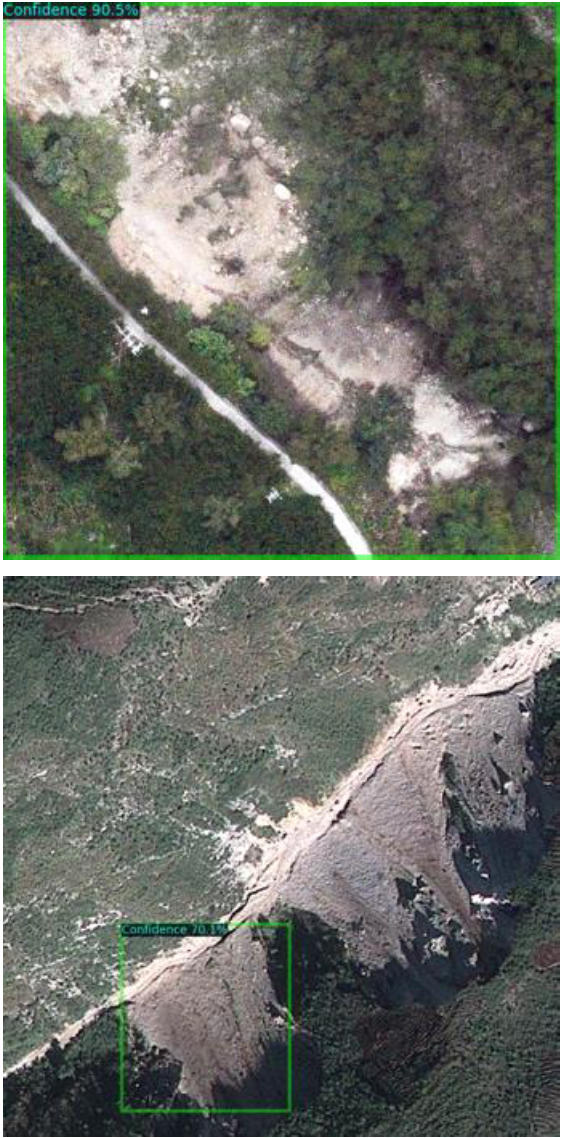}&
\includegraphics[height=1.3in]{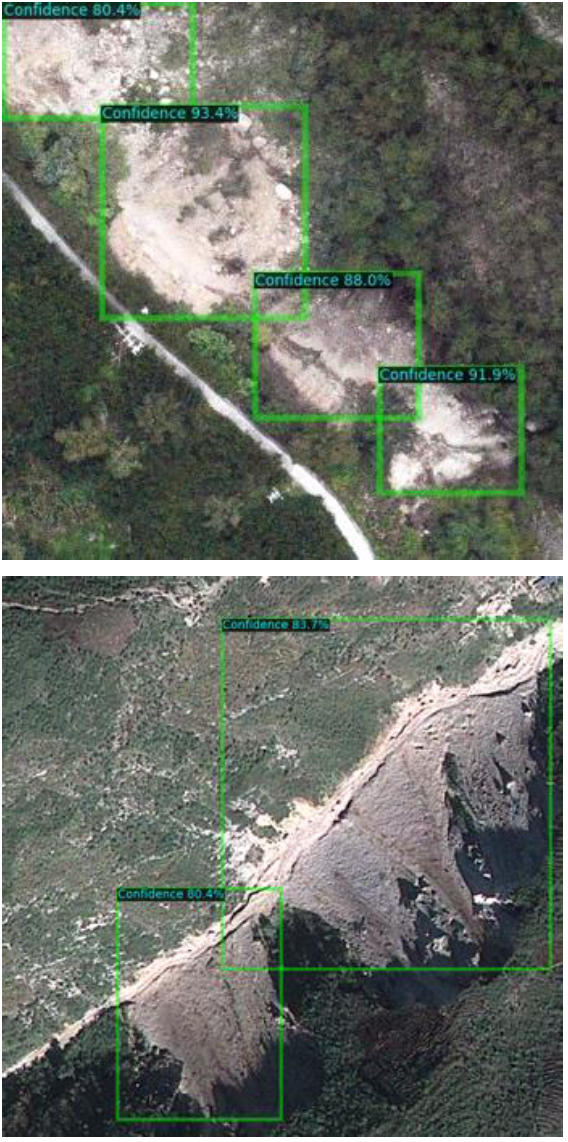}\\
\hspace{0.5cm}DINO &\hspace{0.3cm}YOLOv11&\hspace{0.2cm}DiffusionDet &\hspace{0.4cm}EarthLD \\
\end{tabular}
\caption{Performance comparison of different detection methods in landslide recognition}
\label{Performance_comparison_detection}
\end{figure}

\begin{figure}[t]\scriptsize
\centering
\begin{tabular}{p{1.6cm}p{1.6cm}p{1.6cm}p{1.6cm}}
\includegraphics[height=1.5in]{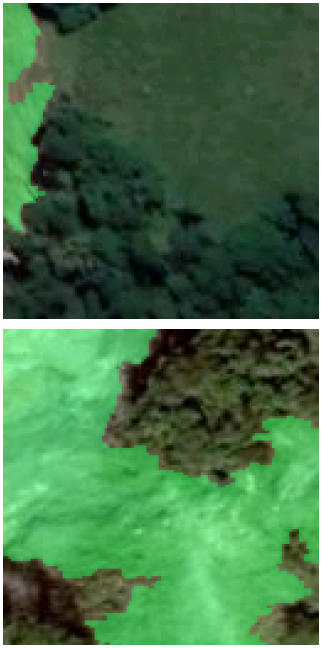}&
\includegraphics[height=1.5in]{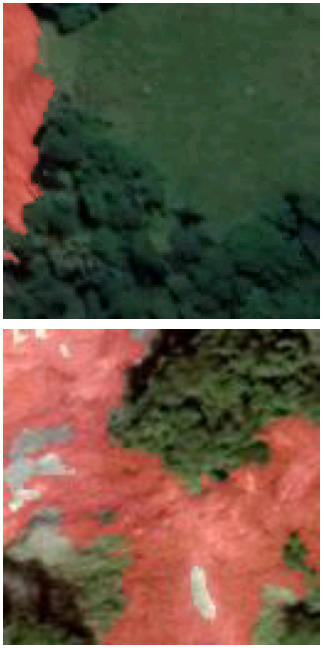}&
\includegraphics[height=1.5in]{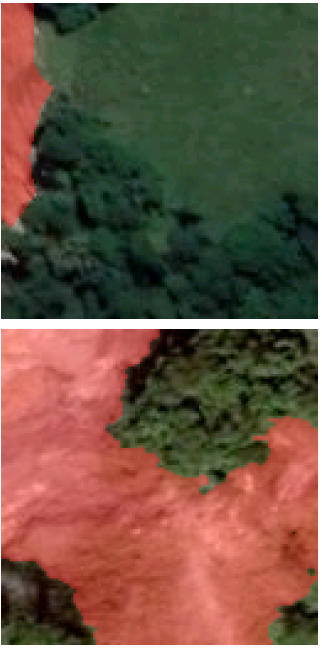}&
\includegraphics[height=1.5in]{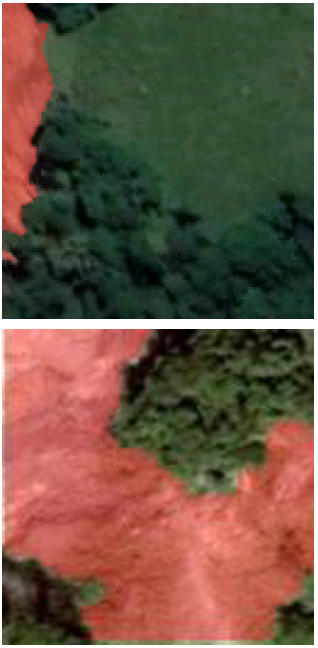}\\
\hspace{0.6cm}GT &\hspace{0.2cm}DeepLabV3+ &\hspace{0.5cm}U-Net++ &\hspace{0.6cm}DFFSA \\
\includegraphics[height=1.5in]{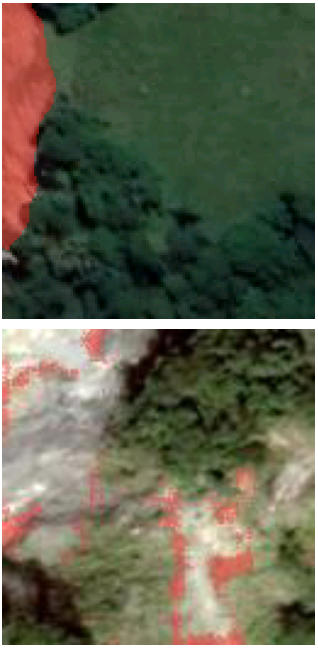}&
\includegraphics[height=1.5in]{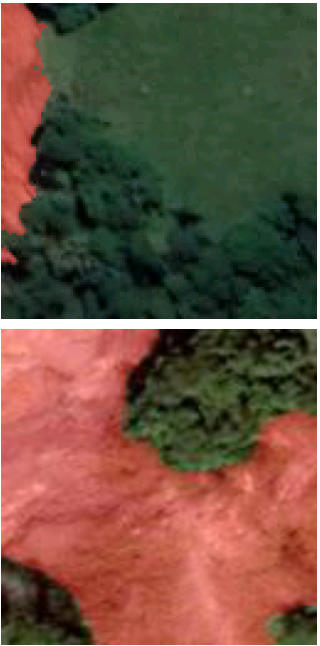}&
\includegraphics[height=1.5in]{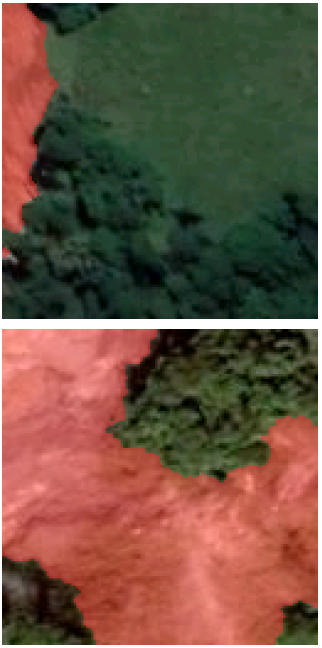}&
\includegraphics[height=1.5in]{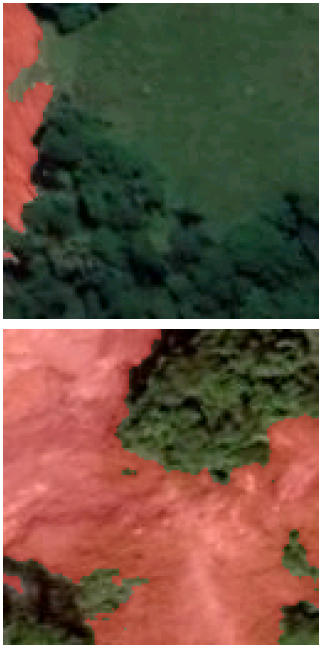}\\
\hspace{0.4cm}SegFormer &\hspace{0.5cm}DINOv2&\hspace{0.1cm}YOLO11s-seg &\hspace{0.5cm}EarthLD \\
\end{tabular}
\caption{Performance comparison of different semantic segmentation approaches in landslide mapping.}
\label{Performance_comparison_segmentation}
\end{figure}

\begin{figure}[t]  
\centering   
\includegraphics[width=1.05in,height=1.03in]{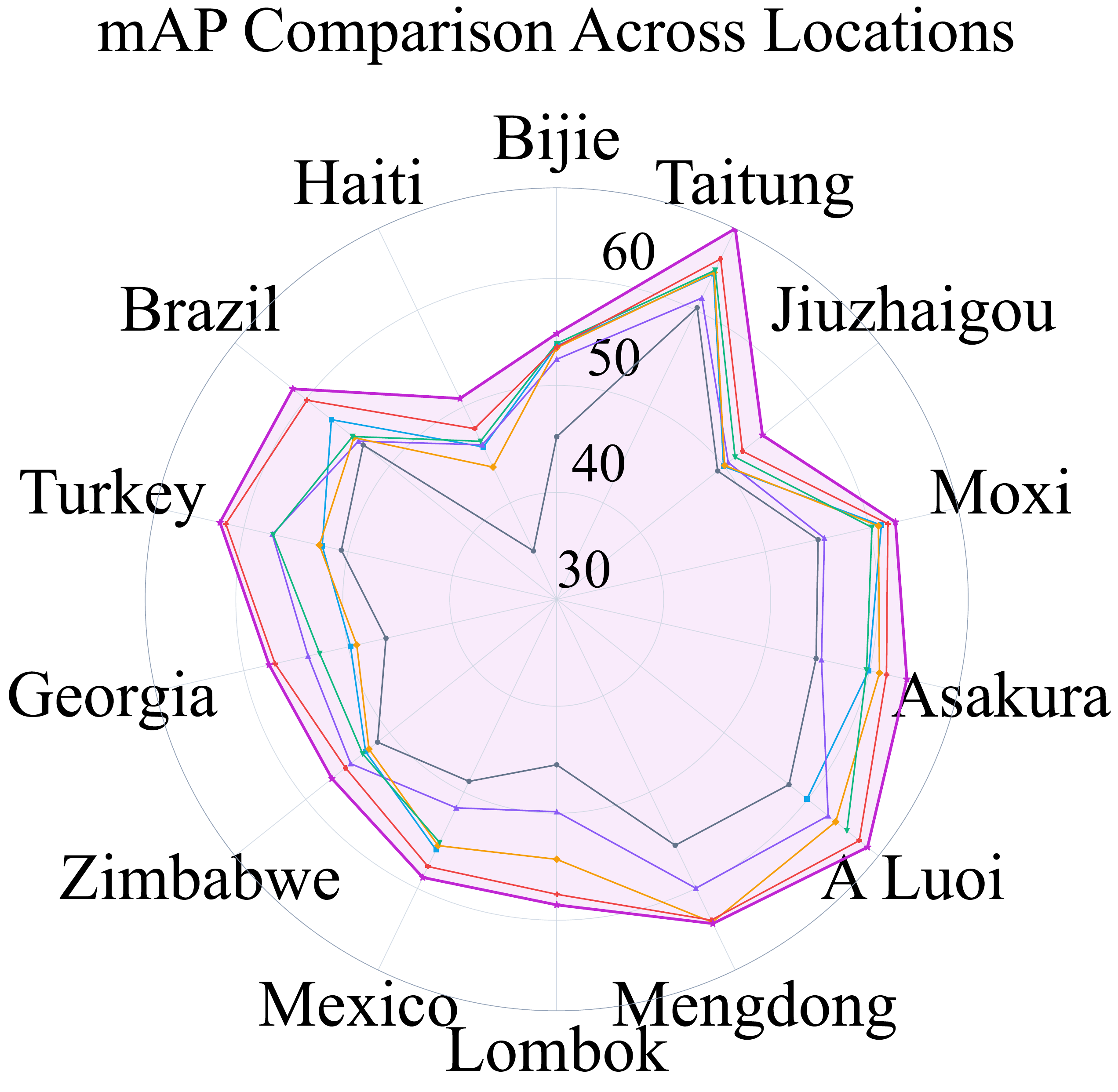}
\includegraphics[width=1.05in,height=1.03in]{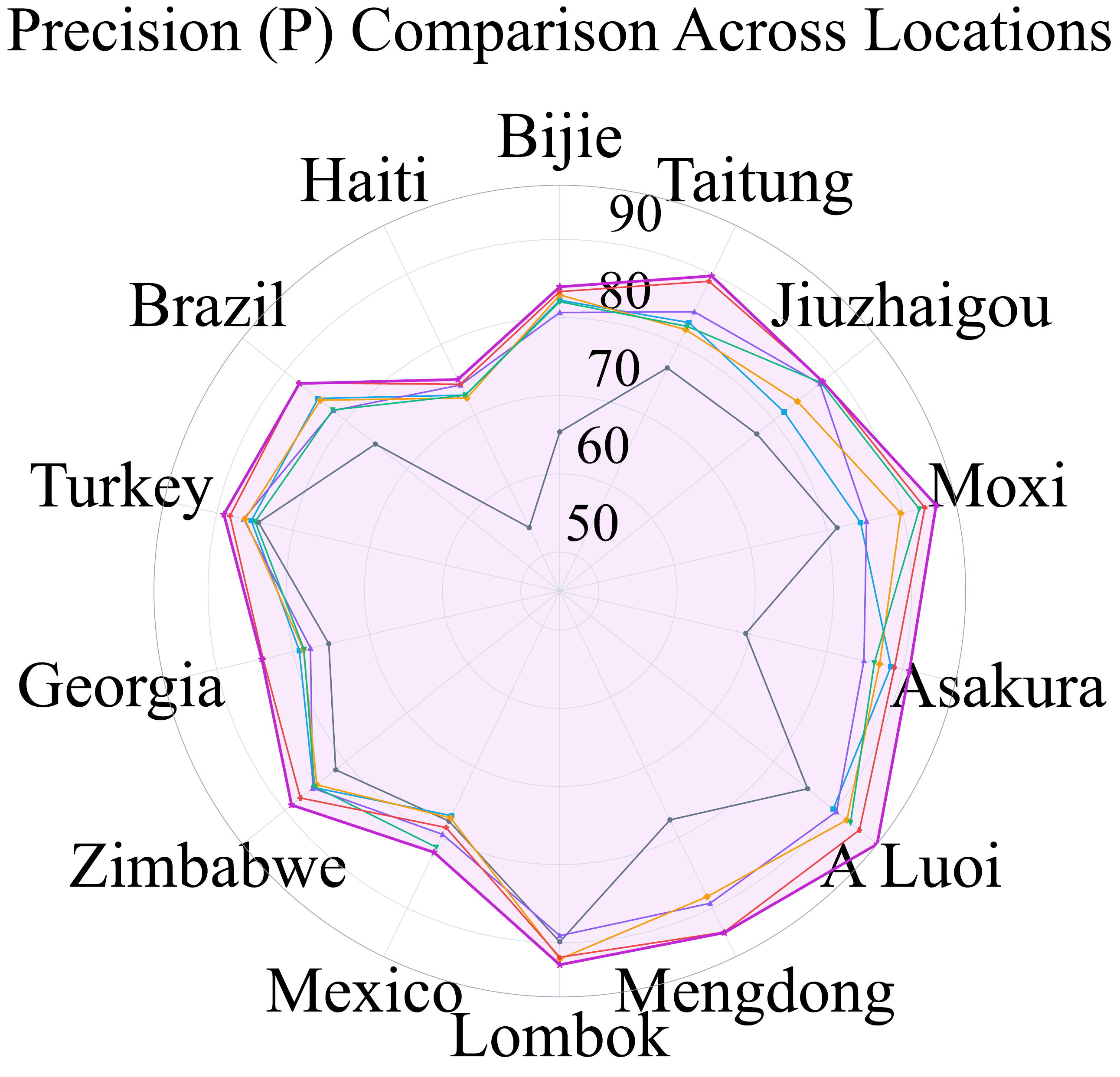}
\includegraphics[width=1.05in,height=1.03in]{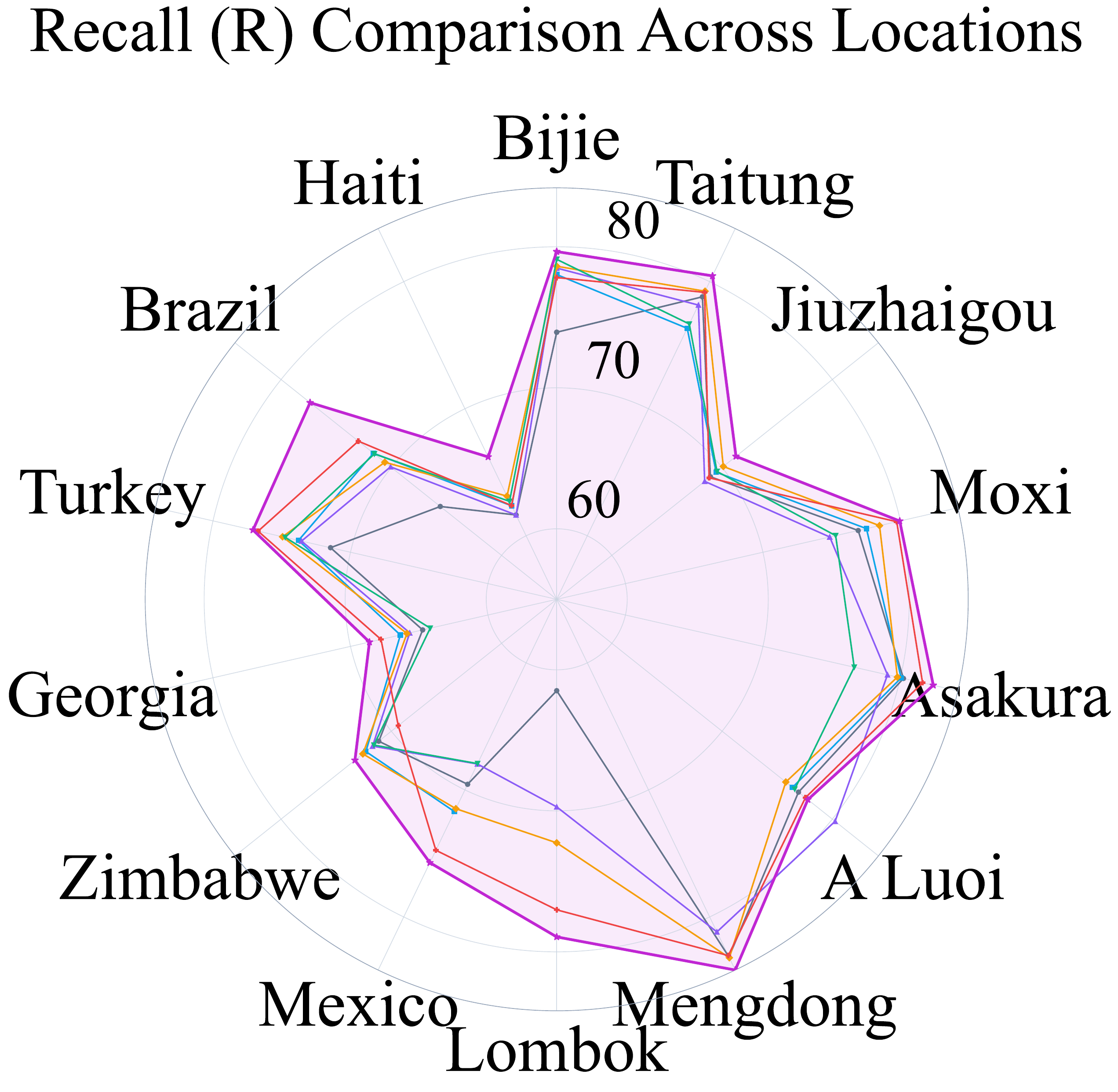}
\includegraphics[width=3.35in,height=0.23in]{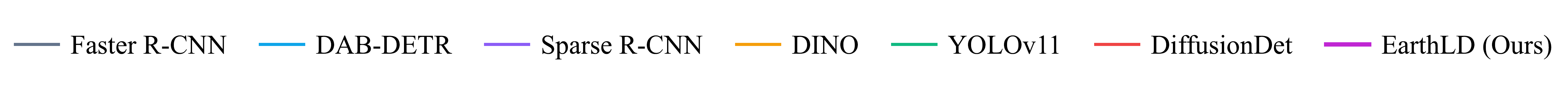}
\caption{Experiments are conducted to evaluate various detection methods using datasets from 10 distinct regions worldwide.}
\label{globalnoise1}
\end{figure}

\begin{figure}[t]\scriptsize
\begin{tabular}{p{1.65cm}p{1.65cm}p{1.65cm}p{1.65cm}}
\includegraphics[height=0.75in]{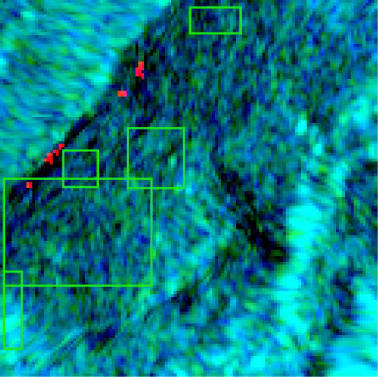}&
\includegraphics[height=0.75in]{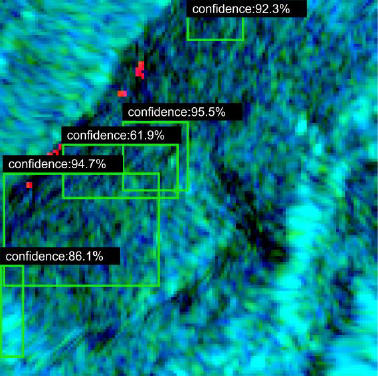}&
\includegraphics[height=0.75in]{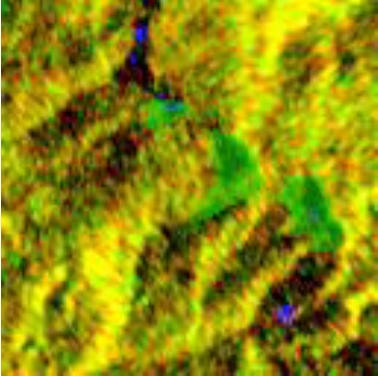}&
\includegraphics[height=0.75in]{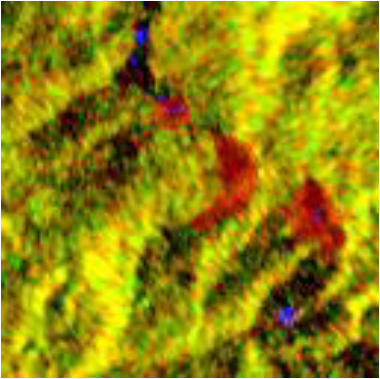}\\
\hspace{0.7cm}GT &\hspace{0.5cm}EarthLD &\hspace{0.7cm}GT &\hspace{0.5cm}EarthLD \\
\end{tabular}
\caption{Landslide detection using SAR data via EarthLD. Left shows a result of landslide identification, and Right displays a result of landslide mapping.}
\label{SAR_data_test}
\end{figure}

\begin{figure}[t]\scriptsize
\begin{tabular}{p{3.75cm}p{3.75cm}}
\includegraphics[height=1.5in]{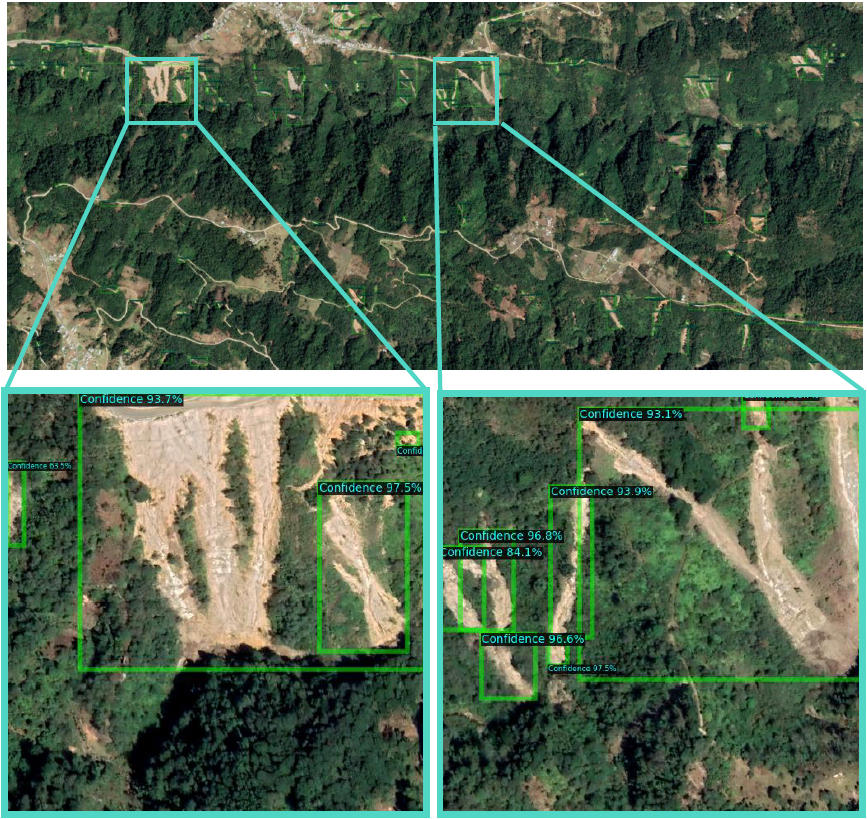}&
\includegraphics[height=1.5in]{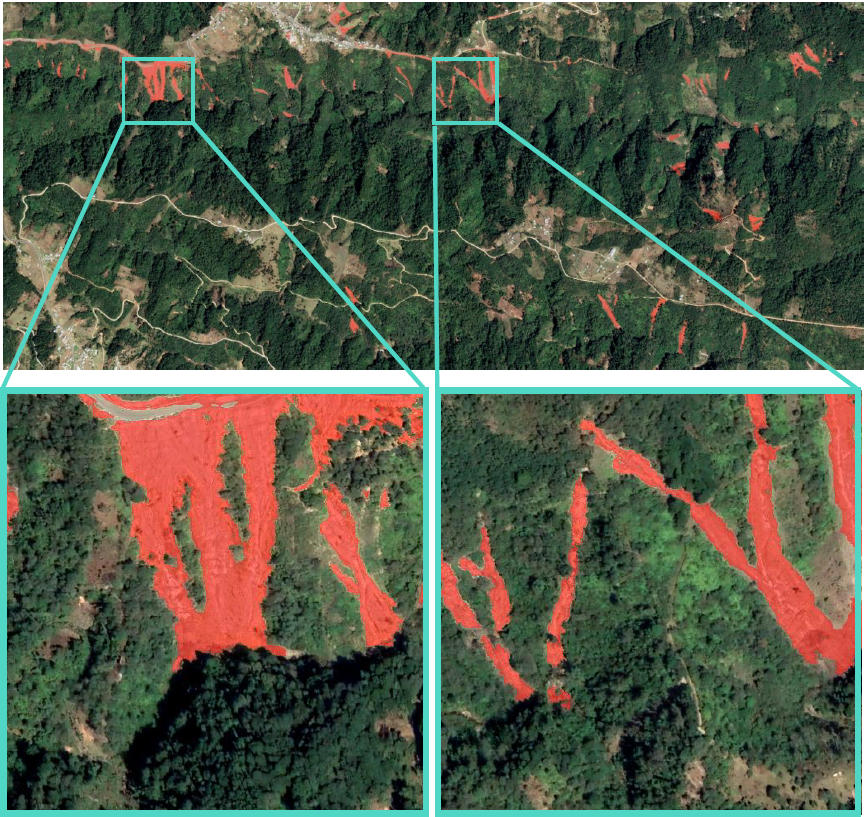}\\
\hspace{0.9cm}Landslide Identification &\hspace{0.9cm}Landslide Mapping \\
\end{tabular}
\caption{EarthLD is applied to landslide detection across large-scale remote sensing imagery.}
\label{large-scale_test}
\end{figure}

\subsection{Unified Learning}
The landslide recognition branch is supervised as an object detector. Its loss combines box variational diffusion, binary landslide objectness classification, and box regression:
\begin{equation}
\mathcal{L}_{\mathrm{det}}
=
\mathcal{L}_{\mathrm{VDM}}^{\mathbf{b}}
+
\lambda_c\mathcal{L}_{\mathrm{focal}}
+
\lambda_1\mathcal{L}_1
+
\lambda_g\mathcal{L}_{\mathrm{GIoU}},
\label{eq:v2-det-loss}
\end{equation}
where $\mathcal{L}_{\mathrm{focal}}$ separates landslide objects from background, $\mathcal{L}_1$ and $\mathcal{L}_{\mathrm{GIoU}}$ supervise box localization, and $\lambda_c,\lambda_1,\lambda_g\in\mathbb{R}^{+}$ are trade-off weights.

The landslide mapping branch is supervised as binary semantic segmentation. Its loss combines mask latent diffusion, Binary Cross-Entropy (BCE), and Dice loss:
\begin{equation}
\mathcal{L}_{\mathrm{map}}
=
\mathcal{L}_{\mathrm{VDM}}^{\mathbf{M}}
+
\lambda_b\mathcal{L}_{\mathrm{BCE}}
+
\lambda_d
\left(
1-
\frac{
2\langle\hat{\mathbf{M}},\mathbf{M}\rangle+\varepsilon
}{
\|\hat{\mathbf{M}}\|_1+\|\mathbf{M}\|_1+\varepsilon
}
\right),
\label{eq:v2-map-loss}
\end{equation}
where
$\langle\hat{\mathbf{M}},\mathbf{M}\rangle
=\sum_{h=1}^{H}\sum_{w=1}^{W}
\hat{\mathbf{M}}_{h,w}\mathbf{M}_{h,w}$,
and $\lambda_b,\lambda_d\in\mathbb{R}^{+}$ are weighting coefficients. BCE performs foreground--background pixel classification, while Dice loss encourages overlap between the predicted and ground-truth landslide regions.

The complete EarthLD objective is
\begin{equation}
\mathcal{L}
=
\mathcal{L}_{\mathrm{det}}
+
\lambda_m\mathcal{L}_{\mathrm{map}}
+
\lambda_{\mathrm{vl}}\mathcal{L}_{\mathrm{NCE}}
+
\lambda_t\mathcal{L}_{\mathrm{trg}},
\label{eq:v2-total}
\end{equation}
where $\mathcal{L}_{\mathrm{NCE}}$ aligns the image embedding $\mathbf{e}_I$ with metadata prompt embeddings $\mathbf{t}_r$, $\mathcal{L}_{\mathrm{trg}}$ supervises trigger estimation, and $\lambda_m,\lambda_{\mathrm{vl}},\lambda_t\in\mathbb{R}^{+}$ balance the objectives. These coefficients may be learned adaptively using GradNorm~\cite{chen2018gradnorm}.

\section{Experiments}
\label{Experiment}

The quantitative analysis employs several evaluation metrics, including Precision (P), Recall (R), and Average Precision (AP). Precision measures the proportion of correctly predicted positive instances or pixel regions, while Recall reflects the proportion of true landslide targets that are successfully detected and segmented.

\subsection{Training Visualization and Ablation Study}
Figure~\ref{Middle_showing} demonstrates the gradual enhancement of the model's feature representation ability for both object detection and semantic segmentation over the course of training, as reflected by the performance at different epochs.

By incorporating pre-trained models trained on expanded datasets, EarthLD achieves stronger feature representations, thereby improving landslide detection and mapping results. Figure~\ref{pre-traintest} illustrates the ablation study regarding CLIP guidance and pre-training. On unseen datasets, the ablated model without CLIP guidance and pre-training yields markedly lower detection accuracy than the full EarthLD.

\subsection{Comparison with Baselines}
Since EarthLD is an all-in-one model capable of simultaneously performing object detection and semantic segmentation, our comparative experiments incorporate two distinct sets of baselines. The first set, established to evaluate object detection performance, includes Faster R-CNN~\cite{ren2016faster}, Sparse R-CNN~\cite{sun2021sparse}, DAB-DETR~\cite{liu2022dabdetr}, DINO~\cite{zhang2022dino}, YOLOv11~\cite{khanam2024yolov11}, and DiffusionDet~\cite{chen2023diffusiondet}. The second set, designed to assess semantic segmentation capability, comprises DeepLabV3+~\cite{chen2018deeplabv3plus}, U-Net++~\cite{zhou2019unetplusplus}, SegFormer~\cite{xie2021segformer}, DINOv2~\cite{oquab2023dinov2}, YOLO11s-seg~\cite{yolo11_ultralytics}, and Seg-Diffusion~\cite{baranchuk2022labelefficient}.

We compare the landslide identification results obtained by different detection methods, as illustrated in Figure~\ref{Performance_comparison_detection}. Similarly, Figure~\ref{Performance_comparison_segmentation} compares the performance of various segmentation methods for landslide mapping. As observed from Figure~\ref{Performance_comparison_detection} and Figure~\ref{Performance_comparison_segmentation}, our EarthLD consistently achieves the best performance in both landslide identification and landslide mapping. Figure~\ref{globalnoise1} presents accuracy evaluations of different detection methods using benchmark datasets from 10 global regions.

\subsection{Multimodal and Large-Scale Imaging Applications}
EarthLD is not only applicable to optical imagery but can also be transferred to SAR data for landslide recognition and detection. As illustrated in Figure~\ref{SAR_data_test}, EarthLD demonstrates effective landslide detection performance on SAR imagery. Furthermore, EarthLD is not limited to small-scale scenes, and it is equally capable of handling landslide detection across large-scale remote sensing images, as shown in Figure~\ref{large-scale_test}. Ultimately, the proposed EarthLD generates a comprehensive landslide detection report, specifying the geographic location, spatial extent, and total count of identified landslide events.

\subsection{Conclusion}
In this paper, we propose EarthLD, an all-in-one framework that unifies landslide recognition, range mapping, trigger estimation, counting, and localization. EarthLD integrates object detection and proposal-guided binary segmentation using VDM and BiFPN for refined denoising and multi-scale feature extraction, complemented by CLIP-derived contextual guidance. Comprehensive evaluations demonstrate EarthLD's superior performance, seamless cross-modal transferability (e.g., optical to SAR imagery), and high efficacy in large-scale remote sensing monitoring and automated statistical reporting.

\bigskip

\bibliography{aaai2027}


\end{document}